\def\year{2022}\relax
\documentclass[letterpaper]{article} 
\usepackage{aaai22}  
\usepackage{times}  
\usepackage{helvet}  
\usepackage{courier}  
\usepackage[hyphens]{url}  
\usepackage{graphicx} 
\def\UrlFont{\rm}  
\usepackage{natbib}  
\usepackage{caption} 
\DeclareCaptionStyle{ruled}{labelfont=normalfont,labelsep=colon,strut=off} 
\usepackage{algorithm}
\usepackage{algorithmic}

\usepackage{threeparttablex} 
\usepackage{multirow}
\usepackage{subcaption}
\usepackage{amssymb}
\usepackage{amsmath}
\usepackage{enumitem}
\usepackage{array}
\usepackage{longtable}
\usepackage{makecell}
\usepackage{microtype}
\usepackage{booktabs}
\usepackage{url}
\usepackage{eurosym}
\usepackage{todonotes}

\newcommand{\figref}[1]{Figure~\ref{fig:#1}}

\newcommand{\tabref}[1]{Table~\ref{tab:#1}}

\newcommand{\model}{\textsc{RDU}}
\newcommand{\sroie}{\textsc{SROIE}}
\newcommand{\inv}{\textsc{INV}}
\newcommand{\bks}{\textsc{BKS}}
\newcommand{\bl}{\textsc{BL}}
\newcommand{\hybrid}{\textsc{Hybrid}}
\newcommand{\single}{\textsc{Single}}

\newcommand{\bert}{BERT$_\text{BASE}$}

\newcommand{\PreserveBackslash}[1]{\let\temp=\\#1\let\\=\temp}
\newcolumntype{C}[1]{>{\PreserveBackslash\centering}p{#1}}
\newcolumntype{R}[1]{>{\PreserveBackslash\raggedleft}p{#1}}
\newcolumntype{L}[1]{>{\PreserveBackslash\raggedright}p{#1}}
 
\newcommand\fone{F\textsubscript{1}}

\usepackage{newfloat}
\usepackage{listings}
\floatstyle{ruled}
\newfloat{listing}{tb}{lst}{}
\floatname{listing}{Listing}
\title{RDU: A Region-based Approach to Form-style Document Understanding}

\author{
    Fengbin Zhu\textsuperscript{\rm 1,2},
    Chao Wang\textsuperscript{\rm 2},
    Wenqiang Lei\textsuperscript{\rm 3},
    Ziyang Liu\textsuperscript{\rm 2},
    Tat Seng Chua\textsuperscript{\rm 1}
}
\affiliations{
    \textsuperscript{\rm 1}
    National University of Singapore,
    \textsuperscript{\rm 2}
    6Estates Pte Ltd,
    \textsuperscript{\rm 3}Sichuan University\\
    \{zhfengbin, wenqianglei\}@gmail.com,\{wangchao, ziyang\}@6estates.com, dcscts@nus.edu.sg

}

\begin{document}

\maketitle

\begin{abstract}
\begin{quote}

Key Information Extraction (KIE) is aimed at extracting structured information (\textit{e.g.} key-value pairs) from form-style documents (\textit{e.g.} invoices),
which makes an important step towards intelligent document understanding.
Previous approaches generally tackle KIE by sequence tagging, which faces difficulty to process non-flatten sequences, 
especially for table-text mixed documents. 
These approaches also suffer from the trouble of predefining a fixed set of labels for each type of documents, as well as the label imbalance issue.
In this work, we assume Optical Character Recognition (OCR) has been applied to input documents, and reformulate the KIE task as a region prediction problem in the two-dimensional (2D) space given a target field. 
Following this new setup, we develop a new KIE model named Region-based Document Understanding (\model) that takes as input the text content and corresponding coordinates of a document, and tries to predict the result by localizing a bounding-box-like region.
Our \model~first applies a layout-aware BERT equipped with a soft layout attention masking and bias mechanism to incorporate layout information into the representations.
Then, a list of candidate regions is generated from the representations via a Region Proposal Module inspired by computer vision models widely applied for object detection.
Finally, a Region Categorization Module and a Region Selection Module are adopted to judge whether a proposed region is valid and select the one with the largest probability from all proposed regions respectively. 
Experiments on four types of form-style documents show that our proposed method can achieve impressive results.
In addition, our \model~model can be trained with different document types seamlessly, which is especially helpful over low-resource documents.

\end{quote}
\end{abstract}

\section{Introduction}
Key Information Extraction (KIE) aims to extract structured information, \textit{e.g.} key (field)-value pairs, from business documents such as invoices, receipts, purchase orders, bank statements, financial reports, \textit{etc.}~\cite{Huang2019SROIE,Jaume2019FUNSD,Gralinski2020Kleister,holt2018extracting,Yu2020PICKPK,wang2021vies}.
In reality, many companies and organisations use this technique to deal with the large amounts of documents that may vary significantly in content and layout, saving much human effort and cost.
In this paper, we address this practical yet challenging task under the assumption that Optical Character Recognition (OCR) has been applied to the documents already.

\begin{figure}[!t]
    \begin{center}
    \includegraphics[scale=0.75]{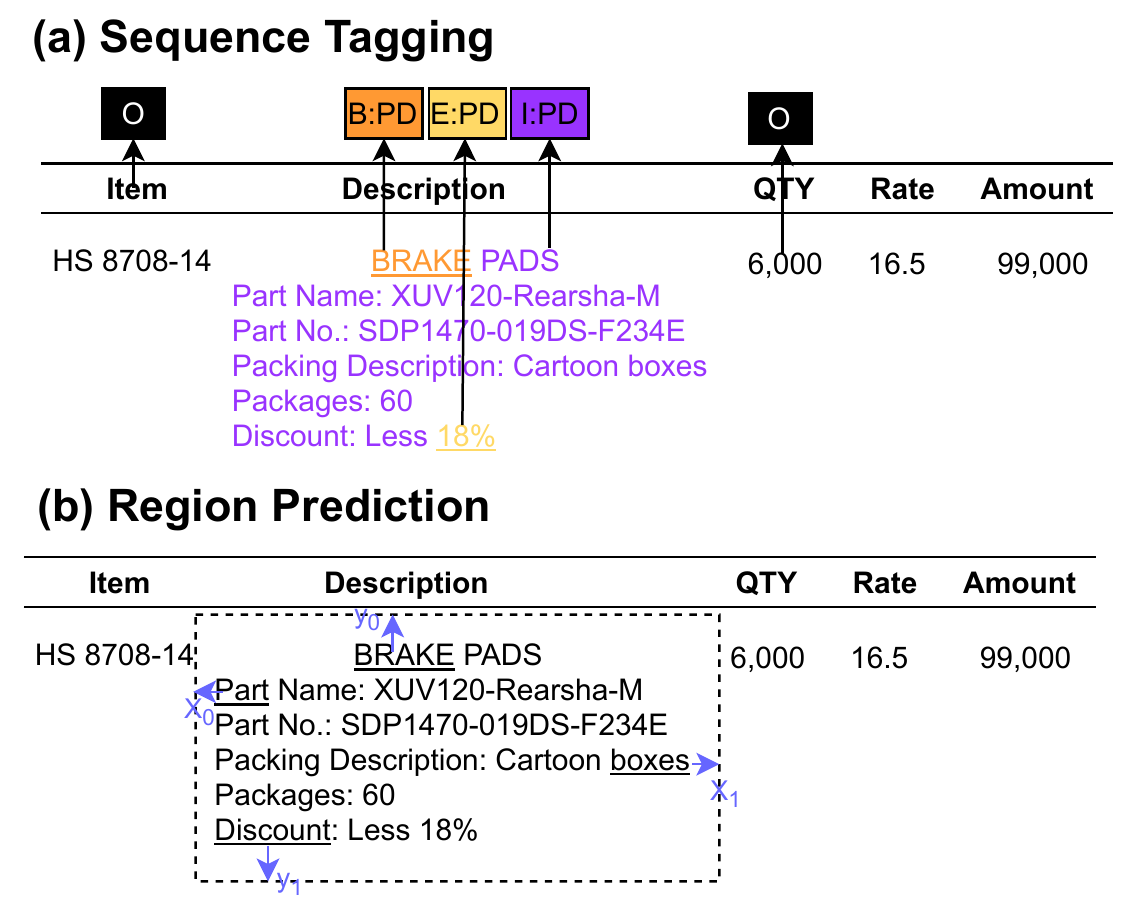}
    \end{center}
    \vspace{-1.5em}
    \caption{ \label{fig:kie}
    Illustration of KIE from documents. 
    Take extracting \emph{Product Description} (PD)  from an invoice as an example. 
    (a) Traditional sequence tagging approach, which assigns each token a predefined label. \textit{B:PD}, \textit{I:PD} and \textit{O:PD} respectively denote the beginning, inside and end of the field \emph{Product Description}.
    The tokens tagged with \textit{O}:outside do not belong to any target field.
    (b) Our region prediction method, \textit{i.e.} locating a bounding-box-like region (in dotted line) by predicting its four boundary tokens (\textit{Part}, \textit{BRAKE}, \textit{boxes}, \textit{Discount}) to generate  the coordinates ($x_0$, $y_0$, $x_1$, $y_1$)  and extracting all the content inside as the result.}
\end{figure}

The majority of previous methods transform the two-dimensional document to a one-dimensional input sequence, and formulate the KIE task as a sequence tagging problem~\cite{hwang2019postocr,xu2020layoutlm,Garncarek2020LAMBERT,hong2021bros,Yu2020PICKPK}.
These methods first serialize the tokens in the document, and then apply an independent tagging model to classify each token to one of the predefined labels, as illustrated in \figref{kie} (a).
However, such a sequence tagging approach suffers from several issues that limit its applicability and performance.  
First of all, due to the diverse and complex spatial layouts of the documents, their serialization is itself a challenging problem that has recently received much research interest~\cite{aggarwal2020form2seq, li2020dice}.
Also, this requires the definition of customized labels for each type of documents, limiting its application to other domains~\cite{Nguyen2021Span}.
Furthermore, sequence tagging may be significantly affected by the existence of imbalanced labels~\cite{Tomanek2009Reducing,li2020dice}.

In this paper, we explore a totally different solution from the traditional sequential tagging.
The key idea is to directly locate a bounding-box-like region on the input document and extract the content inside the region as the KIE result. 
See \figref{kie} (b) for an illustration.  
Given a target field (\textit{Product Description}) and an OCR processed document (an invoice) as input, the goal is to locate a bounding-box-like region by predicting its four boundary tokens to generate its coordinates, which encloses the value of the field (dotted-line rectangle).
This new formulation of the KIE task has several merits.
Firstly, by predicting a region in a two-dimensional space instead of the string(s) in a one-dimensional input sequence, the impact of sequential error can be greatly alleviated. 
Also, by using a two-dimensional region, the KIE model is capable of extracting not only a string of values, but also a column in a table.
This is especially important when dealing with table-text mixed documents, as demonstrated in \figref{kie} (b), where the \emph{Product Description} corresponds to a column of the table.
Moreover, this approach does not require a predefined label set for each document type as in sequence tagging, and hence gets rid of the bad effect due to imbalanced labels.
This also enables a model to be trained on different types of documents seamlessly.

Based on our new formulation of treating KIE as a region prediction problem, we also propose a KIE model named Region-based Document Understanding (\model).
Given an OCR processed document and a target field, \model~first applies a Layout-aware BERT to encode layout information of the document into the representation by leveraging a soft layout attention masking and bias mechanism.
Then, some candidate regions are generated from the representations via a Region Proposal Module, which is inspired by object detection models~\cite{ross2013rcnn,ross2015fast-rcnn,ren2015faster-rcnn}.
Finally, a Region Categorization Module and a Region Selection Module are respectively adopted to judge whether a proposed region is valid, and select the one with the largest probability, from which the final result is extracted.

Note that although we borrow the concept of generating ``region proposals" as in computer vision tasks, we do not tackle KIE as a vision problem; instead, we consider this task from a pure Natural Language Processing perspective.
Our proposed \model~takes as input the text content plus coordinates of a document obtained with OCR, and does not use any visual features, which is computationally efficient. 
We evaluate our method on four different types of documents, including receipts~\cite{Huang2019SROIE}, invoices, bank statements and bill of ladings.
It is shown that \model~can achieve impressive performance on all the types of documents when it is trained separately.
We also find that the performance can be further improved if we combine all four document types to train our model.
This finding indicates that our task formulation and method may be very helpful for KIE over low-resource document types.

\begin{figure*}[!t]
    \begin{center}
    \includegraphics[scale=0.85]{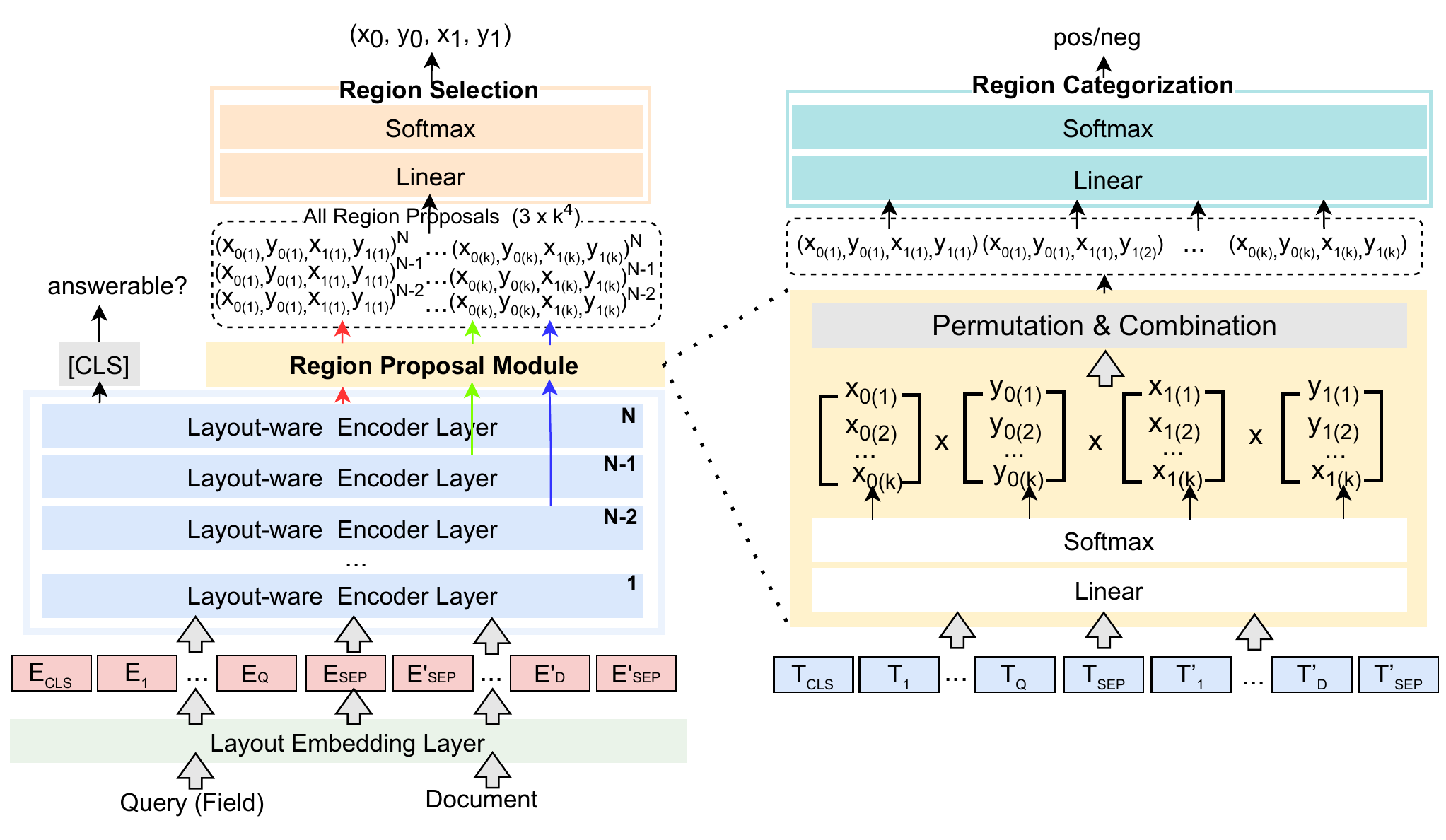}
    \end{center}
    \vspace{-1.5em}
    \caption{ \label{fig:architecture} 
    The architecture of \model.
    $N = 12$ is the number of transformer encoder layers.
    We utilize the outputs from the last three layers to generate the region proposals respectively.
    $k$ is the number of top predictions for each coordinate of a bounding box, and the total number of proposed regions would be $3 * k^4$.
   }
\end{figure*}

\section{Related Work}

\subsection{Key Information Extraction from Documents}
The goal of Key Information Extraction (KIE) is to extract useful structured information from form-style documents such as invoices or medical reports.
Previous approaches to KIE  generally include three groups based on adopted features: text-only, text-layout, and multimodal models.
Text-only methods~\cite{hwang2019postocr,Guo2019EATEN,Nguyen2021Span,zhu2021retrieve} treat the document as plain text and only use text information, which usually offer limited performance due to failure to exploit layout information of the input document.
Multi-modal models~\cite{xu2020layoutlm,xu2021layoutlmv2,xu2021LayoutXLM,Yu2020PICKPK,Appalaraju2021DocFormer,powalski2021going} combine text and vision features and show impressive performance. 
However, they demand processing over raw images, which costs considerable computational resource and time, hampering their application in the real world.
Text-layout models~\cite{katti2018chargrid,Denk2019BERTgrid,Garncarek2020LAMBERT,hong2021bros} adopt both text and layout features that are obtained easily from an Optical Character Recognition (OCR) engine, which largely increases the computational efficiency while keeping comparative performance.
However, most existing text-layout models address KIE  as a sequence tagging problem~\cite{katti2018chargrid,Garncarek2020LAMBERT,hong2021bros,zhu2021tat}. 
This setup would hinder the application of the same model over the variety of document types since sequence tagging requires a predefined label set for each document type and suffers from the issues of imbalanced labels.
In this work,  we re-formalize the KIE task as a region prediction problem in 2D space instead of sequence tagging and propose a \model~model following this new setup, enabling the model to be applied on multiple document types seamlessly.

\subsection{Layout-aware Transformer}
Recently, Layout-aware Transformer has demonstrated noticeable effectiveness for intelligent document understanding, which focuses on exploiting the spatial layout information of the input document.
For example, \cite{Garncarek2020LAMBERT,xu2020layoutlm,hong2021bros,herzig2020tapas} extend the positional embeddings by adding the embedding of layout information;
\cite{powalski2021going,Garncarek2020LAMBERT} choose to modify the raw attention scores in self-attention by adding layout bias terms based on the relative vertical and horizontal distance between two tokens.
In our \model, we adopt both ways to develop a layout-aware BERT.
Firstly, we adapt a similar layout embedding method with LayoutLM~\cite{xu2020layoutlm} to embed the coordinates of the text, but with modifications which is to embed the width and height instead of bottom-right of a coordinate.
We argue that the parameters for the bias terms are not sufficient in number for representing the differences of various layout information~\cite{Garncarek2020LAMBERT,powalski2021going}.
Hence, in our model, in addition to a layout attention bias similar as LAMBERT~\cite{Garncarek2020LAMBERT}, we also develop a soft layout attention masking method to better differentiate the importance of the text according to the spatial relative distance between words.

\section{Methodology}
Instead of tackling KIE by sequence tagging, we reformulate it as region prediction over the input document processed by OCR.
Then following this new formulation, we develop a Region-based Document Understanding model (abbreviated as \model). 
As shown in \figref{architecture},
\model~includes four major parts:
1) a Layout-aware BERT 
to incorporate spatial layout information of the document into the representation;
2) a Region Proposal Module to propose a list of candidate regions;
3) a Region Categorization Module to judge whether a proposed region is valid or not; and
4) a Region Selection Module to select the best one from all proposed regions.

\subsection{Task Formulation: Region Prediction}
Consider a form-style document.
Assume it is pre-processed with an OCR engine and transformed to corresponding text content and coordinates.
Serialize and tokenize the text following a left-to-right and top-to-bottom manner based on the coordinates; each token will have its own coordinates.
Taking as input the serialized text and corresponding coordinates of this document, the task is to locate a bounding-box-like region by predicting its four boundary tokens and extract all the content inside as the final KIE result for a target field.

Formally, we denote the left, top, right and bottom boundary token of the region as $tl$, $tt$, $tr$ and $tb$ respectively.
The coordinates of this region will be $(x^{tl}_{0}, y^{tt}_{0}, x^{tr}_{1}, y^{tb}_{1})$.
To be more clear, we also take \figref{kie} (b) on extracting the given field \emph{Product Description} from an invoice under our new task formulation as an example.
To obtain the target region marked in the dotted-line rectangle, we first predict its four boundary tokens \emph{($tl$:Part, $tt$:BRAKE, $tr$:boxes and $tb$: Discount)}, which are then used to generate the region's coordinates.
Then, all the tokens inside this region will be concatenated following a left-to-right and top-to-bottom manner based on their coordinates as the final result.
For each token $t$ with coordinates $(x^t_0, y^t_0, x^t_1, y^t_1)$ in the document, if $x^{tl}_0 <= x^t_0$, $x^{tr}_1 >= x^t_1$, $y^{tt}_0 <= y^t_0$ and $y^{tb}_1 >= y^t_1$, we then regard this token as inside the target region.

Under this novel formulation, we develop a simple yet effective solution to KIE. Below we elaborate on each part of the proposed \model~model sequentially.  
\subsection{Layout-aware BERT}

The proposed \model~first applies a Layout-aware BERT built on top of the vanilla \bert~\cite{Devlin2018Bert} to generate the representations.
The vanilla \bert~\cite{Devlin2018Bert} is trained on plain text, and hence is ``unaware'' of the layout information of the document, which however is very crucial to KIE over form-style documents. 
To effectively encode spatial layout information from the input document into BERT representations, we make two modifications.
First, we directly embed the coordinates of each token and add them to the input embeddings in the embedding layer.
Second, we apply a soft layout attention masking and bias mechanism when computing the attention scores in self-attention, which depends on the horizontal and vertical relative distance between the tokens.
More details are given below. 

\subsubsection{Layout-aware Embedding}
\begin{figure}[htb]
    \begin{center}
    \includegraphics[scale=0.6]{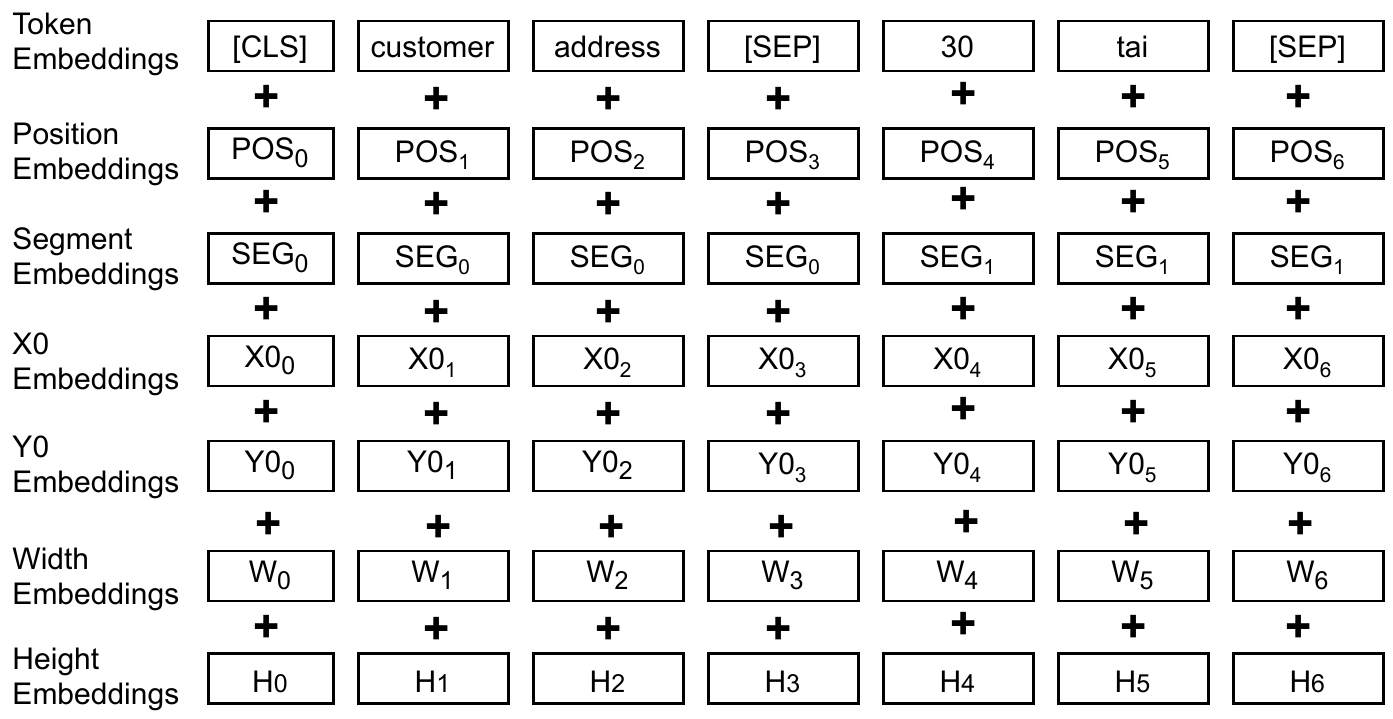}
    \end{center}
    \vspace{-1.5em}
    \caption{\label{fig:emb} 
    The embedding layer of \model~model. Layout embedding is directly added with input embedding to encode layout information.  
   }
\end{figure}

As explained in our new formulation, each token in the input sequence has its own coordinates obtained from OCR processing.
We embed the coordinates and add them to the input embedding for this token similar as in LayoutLM~\cite{xu2020layoutlm}, so as to encode the layout information.
See \figref{emb} for an illustration.
In particular, we normalize and discretize the coordinates of the document to integers within [0, 700] for width and [0, 1000] for height, basically following an A4 paper width/height ratio.
Then, we use four separate embedding layers to encode the \emph{$x_0$},  \emph{$y_0$}, \emph{Width} and \emph{Height} of the coordinates of a token.
Lastly, we add the four layout embeddings to the input embeddings.

\subsubsection{Layout-aware Encoder Layer}
We extend the typical transformer encoder layer by applying a soft layout attention masking and bias mechanism to the raw attention scores.
This mechanism introduces a soft masking term $w_{ij}$ and a layout attention bias term $b_{ij}$ into the computation of the raw attention scores $\alpha_{ij}$.
These two terms reflect the relative distance between two tokens $i$ and $j$ in a two-dimensional space based on the intuition that two spatially close tokens should be strongly correlated in semantics and vice versa.

To obtain the final raw attention scores $\alpha'_{ij}$, we multiply the raw attention scores $\alpha_{ij}$ with the masking terms $w_{ij}$ first and then add the bias terms $b_{ij}$.
Formally, 
\begin{align}
  \alpha'_{ij} &=  w_{ij} \times \alpha_{ij}  + b_{ij}.
\end{align}

To be more specific, for a token $i$ with its coordinates denoted as $r_i$ = ($x_0$, $y_0$, $x_1$, $y_1$), we first compute the center point: $C(x_i, y_i)$ = ($(x_0 + x_1) / 2$,  $(y_0 + y_1) / 2$); similarly we compute the center of a token $j$ : $C(x_j, y_j)$.

To compute the final masking term $w_{ij}$, we first enable the token $i$ to attend to the token $j$ by first computing the horizontal masking term $hw_{ij}$ and the vertical masking term  $vw_{ij}$.
We then obtain the final masking term $w_{ij}$ by averaging the above two masking terms.
Formally, 
\begin{align}
  hw_{ij} &=  \sigma( \gamma \times (h_{\theta} - h_{ij})) \\
  vw_{ij} &=  \sigma( \gamma \times (v_{\theta} - v_{ij})) \\
  w_{ij} &= (hw_{ij} + vw_{ij}) / 2. 
\end{align}
Here
\(\sigma\) indicates the \emph{sigmoid} function;
$h_{ij}$, $v_{ij}$ respectively denote the horizontal and vertical distance between tokens $i$ and $j$, which are computed by $\mid x_i-x_j \mid$ and $\mid y_i-y_j \mid$ respectively;
$h_{\theta}$ and $v_{\theta}$ are the thresholds to make the tokens within the horizontal and vertical distances more important than the ones outside;
$\gamma$ is a steepness coefficient of the function $\sigma$;
$h_{\theta}$, $v_{\theta}$ and $\gamma$ are hyper parameters.
Since the query/field does not have layout information, we still use the global attention similar to the standard transformer~\cite{Ashish2017attention}, meaning the tokens in the query can attend to or be attended by all the tokens in the document.

Adding a layout attention bias term $b_{ij}$ is previously adopted in \cite{Garncarek2020LAMBERT,powalski2021going}.
In this work, we follow the same approach:
\begin{align}
  b_{ij} &= H(h_{ij}) + V(v_{ij}) 
\end{align}
where \(H(h)\) and \(V(v)\) are trainable weights defined for every
integer \(h \in [0, 700]\) and \(v \in [0, 1000]\).

\subsection{Region Proposal Module}
Our Region Proposal Module aims to propose a list of regions that likely include the target value in a 2D space, which is inspired by object detection models~\cite{ross2013rcnn,ross2015fast-rcnn,ren2015faster-rcnn} but avoids using raw images.

A region in our setup is decided by four boundary tokens represented by ($x^r_0$, $y^r_0$, $x^r_1$, $y^r_1$) corresponding to left, top, right and bottom boundaries respectively.
We take the computation of coordinate $x^r_0$ as an example.
As illustrated on the right in~\figref{architecture}, this module takes the representations from the layout-aware BERT as input, and predicts the top $k$ candidate tokens first using a softmax function. Formally,
\begin{align}
    l_{x^r_0} &= W_1 \cdot T_n \\
    p_{x^r_0} &= softmax(l_{x^r_0} )\\
    c_{x^r_0} &= argsort(p_{x^r_0})[:k].
\end{align}
Here $W_1$ denotes trainable variables; $T_n$ stands for the outputs of the $n$th layer of the layout-aware BERT;
$l_{x^r_0}$ and $p_{x^r_0}$ are the logits and probabilities of the tokens; 
$c_{x^r_0}$ denotes the predicted top $k$ candidate tokens for $x_0$;
$\cdot$ is dot product.

Similarly, we can get the top $k$ candidate tokens for $y^r_0$, $x^r_1$ and $y^r_1$.
After that, a permutation and combination process is applied on all candidate tokens to generate $k^4$ region proposals.
In practice, we use the last three layers of layout-aware BERT (based on empirical validation) to generate region proposals, resulting in a total of $3 * k^4$ proposals.
Among these proposals, we mask the invalid ones if $x_0 > x_1$ or $y_0 > y_1$.

\subsection{Region Categorization Module}
For each proposed region, we assign a binary class label to define it as valid or not.
Similar to~\cite{ren2015faster-rcnn}, we assign a positive label to
(i) the region with the highest Intersection-over-Union (IoU) overlap with the ground-truth box, or (ii) the region that has an IoU overlap higher than $0.7$ with the ground-truth bounding box.
We assign a negative label to a non-positive region if its IoU ratio is lower than $0.1$.
We mask the regions that are neither positive nor negative to ensure that they do not contribute to the training objective.

Before categorizing each region, we need to compute its representation first.
Specifically, we represent a region with two features: 
(i) a content feature that is obtained by applying max-pooling over all tokens within the region, and (ii) a feature of boundaries using the logits of the four boundary tokens obtained in the region proposal step.
The two features go through a feed-forward layer separately to get the same hidden size $h = 64$, which are denoted as $r_{c}$ and $r_{b}$,  and then are concatenated to form the representation of the region denoted as $r_{region}$.
Formally,
\begin{align}
    r_{c} &= W_2 \cdot max(T_r)\\
    r_{b} &= W_3 \cdot [l_{x^r_0};l_{y^r_0};l_{x^r_1};l_{y^r_1}]  \\
    r_{region} &= [r_{c};r_{b}]
\end{align}
where $W_2\in\mathbf{R^{h \times d}}$ and $W_3 \in \mathbf{R^{h \times 4}}$ are trainable variables;
$T_r$ are all tokens within the region;
$max$ is max-pooling;
$;$ means concatenation;
$d$ is the embedding size.
After that, a linear layer and a softmax layer are used to predict whether the region is positive or negative:
\begin{equation}
    p_{rc} = argmax(softmax(W_4 \cdot r_{region}))
\end{equation}
where $W_4$ is a trainable variable.

\subsection{Region Selection Module}
We the describe how we select the best region from a list of region proposals obtained in the previous step.
We treat the region that has the highest IoU overlap with the ground-truth region as the target region.
Firstly, we use the same method as in Region Categorization Module to compute the representation of each region, and then apply a linear layer and a softmax layer on all proposed regions.
The region with the highest probability is selected by
\begin{align}
    p_{rs} &= argmax(softmax(W_5 \cdot r_{region}))
\end{align}
where $W_5$ is a trainable variable.

\subsection{Unanswerable Queries}
The unanswerable queries refer to the cases where the value of the target field does not exist in the given document. 
These cases are not rare. 
To tackle them, we follow the traditional setting of \bert~which uses the representation of the token {\tt [CLS]} for a binary classification task.

\subsection{Loss Function}
Assume the region that has the highest IOU overlap with the ground-truth region $\mathbf{G}^\textrm{region}$ is  $\mathbf{R}^\textrm{region}$.
The loss function is defined as the sum of answerable loss $\mathcal{L}_\textrm{answerable}$, region categorization loss $\mathcal{L}_\textrm{rc}$ and region selection loss $\mathcal{L}_\textrm{rs}$, which are computed as 
\begin{align}
    \mathcal{L}_\textrm{answerable} &=  \textrm{NLL}(\mathbf{P}^\textrm{answerable}, \mathbf{G}^\textrm{answerable}) \\
    \mathcal{L}_\textrm{rc} &=  \textrm{NLL}(\mathbf{P}^\textrm{cat}, \mathbf{G}^\textrm{cat}) \\
    \mathcal{L}_\textrm{rs} &=  \textrm{NLL}(\mathbf{P}^\textrm{region}, \mathbf{R}^\textrm{region}) \\
    \mathcal{L} &=  \mathcal{L}_\textrm{answerable} + \alpha  \mathcal{L}_\textrm{rc} + \beta  \mathcal{L}_\textrm{rs}.
\end{align} 
Here NLL(·) is the negative log-likelihood loss; 
$\mathbf{G}^\textrm{answerable}$ is the ground-truth indicating whether the value of the target field exists in the document or not;
$\mathbf{G}^\textrm{cat}$ is the ground-truth category (\textit{i.e.} positive or negative) of a region;
$\mathbf{P}^\textrm{answerable}$, $\mathbf{P}^\textrm{cat}$ and $\mathbf{P}^\textrm{region}$ are the log-probability of the answerable prediction, region category prediction and region selection respectively;
$\alpha$ and $\beta$ are the loss weights.

The distribution of proposed regions is usually unbalanced, with a bias towards negative samples.
To eliminate the impact of the bias upon training, we set a dynamic loss weight by computing the percentage of the opposite label (positive vs. negative) over the sum of the two labels and multiplying this percentage with the corresponding loss of this label.

\section{Experiments}

\subsection{Datasets}
We conduct experiments on four KIE datasets, as summarized in \tabref{kie-dataset}.
Among them, \sroie~\cite{Huang2019SROIE} is the most widely used public dataset for KIE, which releases $626$ receipts for training and hides $347$ receipts for testing.
We further split the $626$ receipts to train dev and test set in our experiments.
Since \sroie~also offers its OCR results, including the text and corresponding coordinates, we can obtain the bounding box of the answers via pre-processing.

\begin{table}[htb]
\centering
\footnotesize
\begin{tabular}{L{1cm}L{2cm}R{1.0cm}R{1cm}R{1cm}}
\toprule

\bf No. & \bf Dataset  & \bf Train & \bf Dev & \bf Test \\
\midrule

1 & \sroie & 439 & 62 & 125 \\
2 & \inv  & 1,085 & 155 & 310 \\
3 & \bks  & 614 & 87 & 175 \\
4 & \bl & 568 & 80 & 161 \\

\bottomrule
\end{tabular}
\caption{KIE datasets used for evaluating model  performance.
}
 \label{tab:kie-dataset}    
\end{table}

\begin{table}[htb]
\centering
\footnotesize
\begin{tabular}{L{1.3cm}C{1.2cm}C{1.2cm}C{1.2cm}C{1.2cm}}
\toprule
\multirow{2}{*}{ \bf Model } & \bf \sroie & \bf \inv & \bf \bks & \bf \bl \\
\cmidrule{2-5}
& \bf EM/\fone{} & \bf EM/\fone{} & \bf EM/\fone{} & \bf EM/\fone{} \\
\midrule
BERT (S) & 55.71/63.43 & 66.62/79.57 & 67.24/71.14 & 27.00/34.72 \\
BERT (H) & 56.31/64.54 & 66.62/80.22 & 68.76/72.88 & 33.12/40.24 \\
\midrule
\model~(S) & 89.18/93.85 & 75.34/88.56 & \bf 87.79/91.58 &  \bf 75.94/82.15 \\
\model~(H) & \bf 90.98/95.32 & \bf 77.96/90.68 & 87.17/90.70 & 74.80/81.68 \\

\bottomrule
\end{tabular}
\caption{Performance comparison on the test sets of four datasets.
(S) means \single~model that is trained sorely on each document type; (H) means \hybrid~model that is trained on the combination of four document types.}
 \label{tab:main-result}    
\end{table}

\begin{table}[htb]
\centering
\footnotesize
\begin{tabular}{L{0.8cm}L{3.1cm}C{0.5cm}C{0.5cm}C{0.5cm}C{0.5cm}}
\toprule
\multirow{2}{*}{ \bf Dataset } & \multirow{2}{*}{ \bf Field } & \multicolumn{2}{c}{ \model~(S)} & 
\multicolumn{2}{c}{ \model~(H)} \\
\cmidrule{3-6}
& & \bf EM & \bf \fone{} & \bf EM & \bf \fone{} \\
\midrule
\multirow{4}{*}{ \sroie } & Company &  92.00 & \bf 96.85 & 92.00 & 96.68 \\
& Date &  98.40 & 98.40 & 99.20 & \bf 99.20 \\
& Address &  80.65 & 94.55 & 83.87 & \bf 96.62 \\
& Total &  85.60 & 85.60 & 88.80 & \bf 88.80 \\
\midrule
\multirow{8}{*}{ \inv } & Invoice No. &  86.77 & \bf 87.17 & 86.45 & 87.07 \\
& Issuance Date & 95.16 & 95.38 & 99.35 & \bf 99.35 \\
& Customer Name &  86.60 & \bf 96.89 & 86.60 & 96.85 \\
& Customer Address &  71.91 & \bf 93.68 & 72.24 & 93.53 \\
& Vendor Name & 69.16 & 86.66 & 72.08 & \bf 89.03 \\
& Total Amount &  83.11 & 83.11 & 84.77 & \bf 84.77 \\
& Product Code* & 52.45 & 68.38 & 58.74 & \bf 76.47 \\
& Product Description* & 42.35 & 86.26 & 50.53 & \bf 90.70 \\
\midrule
\multirow{7}{*}{ \bks } & Bank Name &  87.38 & \bf 89.48 & 87.70 & 89.10 \\
& Statement Date &  87.77 & \bf 95.67 & 87.46 & 94.70 \\
& Account Number &  85.60 & \bf 86.12 & 84.44 & 85.21 \\
& Account Holder Name &  93.08 & \bf 97.12 & 91.51 & 96.17 \\
& Account Holder Address &  82.93 & \bf 96.32 & 83.74 & 95.54 \\
& Opening Balance &  91.88 & 91.88 & 92.19 & \bf 92.19 \\
& Closing Balance &  84.38 & \bf 84.38 & 81.88 & 81.88 \\
\midrule
\multirow{8}{*}{ \bl } & BL No. &  75.78 & 78.57 & 79.50 & \bf 80.64 \\
& Issuance Date &  70.00 & \bf 74.09 & 68.75 & 72.04 \\
& Issuance Place &  72.30 & \bf 79.29 & 73.65 & 78.15 \\
& Vessel Name &  83.75 & \bf 88.83 & 81.25 & 88.56 \\
& Voyage No. &  77.92 & \bf 79.65 & 76.62 & 78.46 \\
& Port of Loading &  77.02 & 86.11 & 75.16 & \bf 86.77 \\
& Port of Discharge &  81.76 & \bf 88.66 & 78.62 & 88.37 \\
& Place of Delivery &  66.67 & \bf 81.63 & 61.79 & 79.62 \\
\bottomrule
\end{tabular}
\caption{
Field-level performance on the test sets of four datasets. 
The field name with * means ``column field'' whose target value is a column of a table. 
\textit{Single} means  it is trained sorely on one document type; \textit{Hybrid} means it is trained on all document types.
The best \fone{} score is \textbf{highlighted}.}
 \label{tab:field-result}    
\end{table}

The other three datasets, \textit{i.e.} \inv~(Invoices), \bks~(Bank Statements) and \bl~(Bill of Ladings), are built and annotated by ourselves.
We first gather invoices, bank statements and bill of ladings in English from a variety of real world companies, and process these data with Google OCR\footnote{https://cloud.google.com/vision/docs/pdf} for scanned PDFs or images and Apache PDFBox\footnote{https://pdfbox.apache.org/} for readable PDFs.
Then, we employ about $20$ university students majored in finance or equivalent disciplines as annotators to annotate each field value by drawing a rectangle on the document with our self-developed dashboard. 
Each of the datasets is split into train, dev and test set following the ratio \emph{7:1:2}.

\subsection{Evaluation Metrics}
We adopt two popular evaluation metrics for extractive question answering (QA) area: Exact Match (EM) and numeracy-focused \fone{} score similar to those used in  \citep{Dua2019DROP},  
which both measure the overlap between a bag-of-words representation of the golden answers and predictions.
Note that the numeracy-focused \fone{} will be 0 in case of a number mismatch between gold and predicted answers regardless of any other word overlap.

\subsection{Baseline}
We adapt the classic vanilla \bert~to our new task setting as a baseline.
After serializing and tokenizing the text following a left-to-right and top-to-bottom manner, vanilla BERT takes the serialized tokens as input and generates the representations. 
Then, we extend it to predict the four boundary tokens of a region similarly to its basic approach of predicting the start and end tokens of a span answer in extractive QA~\cite{Devlin2018Bert}.
For each token in the input sequence, its probability scores of being the left/top/right/bottom boundary token are computed by a softmax over all the input tokens.
The score of each candidate region is computed by multiplying the probability scores of its four boundary tokens and the valid region with the maximum score is selected.

\subsection{Implementation}
Our \model~is initialized with vanilla \bert~and pretrained on form-style documents in an unsupervised manner, since vanilla \bert~lacks the understanding of the document layout information.
Our pre-training data include text and corresponding bounding box information for 5.3 million English pages of financial statements, invoices, receipts, bank statements and bill of ladings.
For the readable PDFs, we apply Apache PDFBox library to extract the text and corresponding coordinates;
for scanned PDFs or images, we use Google OCR.
We only use the Masked Language Model~(MLM) training objective, set longer sequence length and dynamically mask training data for pre-training, following the practice in RoBERTa~\cite{Liu2019RoBERTa}.

For all the models, the hidden size and embedding size $d$ is set to $768$, and sequence length is $1024$.
We train all models with an Adam optimizer~\cite{Diederik2015Adam} and a learning rate of $3e-5$.
For Layout-aware BERT, the thresholds $h_{\theta}$ and $v_{\theta}$ are set to $300$ and $500$.
The steepness coefficient $\gamma$ of the sigmoid function is $1.0$.
In Region Proposal module, we set $k$ as 3, \textit{i.e.} taking the top 3 tokens as candidates for each coordinate.
We set loss weights $\alpha$, $\beta$ to $1.0$ and $0.01$ respectively.
In pre-training, we train our model using two v100 GPUs for $3$ weeks, and in fine-tuning we use one v100 GPU to train one model for up to 30 epochs in $2$ days.

\subsection{Comparison Results}
To evaluate the effectiveness of our approach, we run our \model~and the baseline method that are trained separately on each document type, namely \inv~(Invoices), \sroie~(Receipts), \bks~(Bank statements) and \bl~(Bill of ladings), as well as a hybrid of all of them.
Models trained on a single document type are marked with a `S' in bracket, meaning \single; similarly, models trained on combined four types of documents are marked with `H' for \hybrid.
We summarize the results in \tabref{main-result} and 
present the performance of our \model~ model for each field in \tabref{field-result}.

\begin{figure*}[!t]
    \begin{center}
    \includegraphics[scale=0.84]{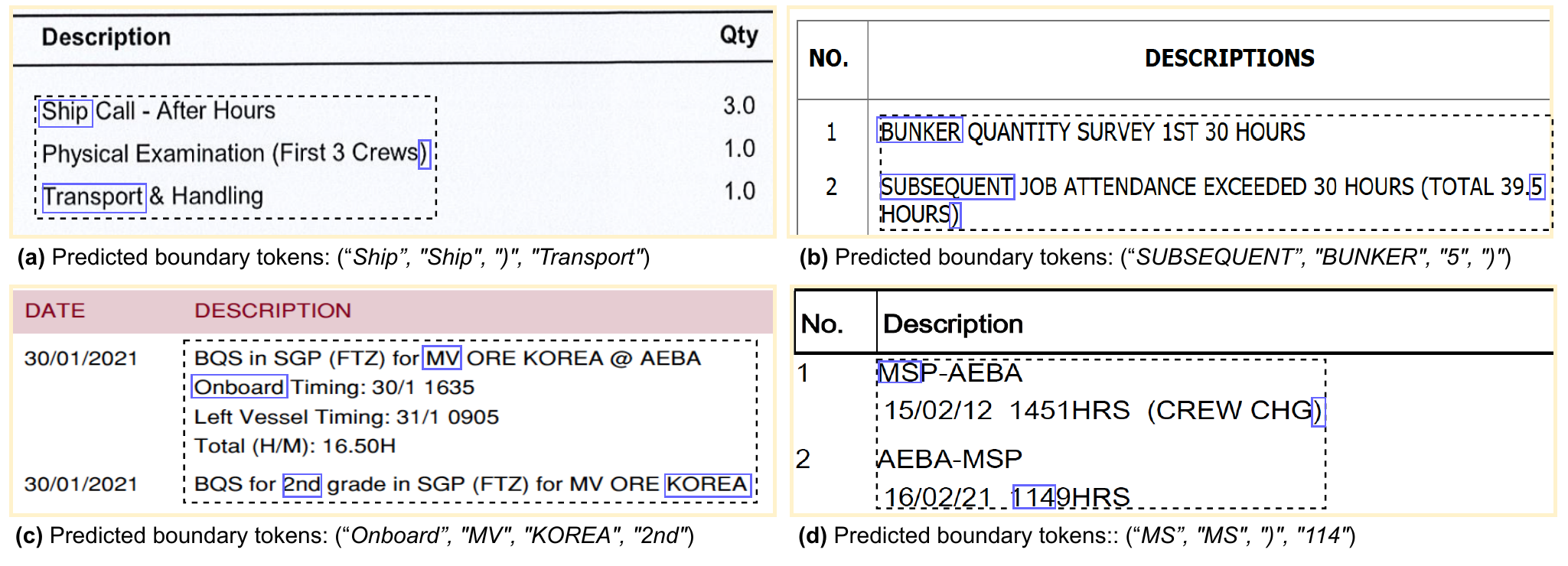}
    \end{center}
    \vspace{-1.5em}
    \caption{ \label{fig:col}
    Some examples for extracting a column field \emph{Product Description} (PD)  from invoices with \model~(\hybrid) model.
    The predicted region is marked in the dark dotted  rectangle and the predicted four boundary tokens are marked in the blue rectangle.
    The predicted boundary tokens are in the order of ($x_0$, $y_0$, $x_1$, $y_1$).} 
\end{figure*}

It can be seen that when trained on each type of documents separately,  our \model~model performs much better than the baseline, demonstrating the effectiveness of our design. 
Especially on \bl, the proposed \model~can gain over 47.0\% \fone{} score higher than the baseline method.

We further train both baseline and \model~on the combination of four types of documents, \textit{i.e.} BERT~(H) and \model~(H). 
From \tabref{main-result}, we find that the \hybrid~models perform similarly or even better than the \single~models.
In particular, BERT (H) outperforms the BERT (S) on each document type.
And, \model~(H)~performs better on two document types i.e., \inv~and \sroie~than \model~(S), gaining 1.47\% on \sroie~and 2.12\% on \inv~in \fone{} score.
This may be because receipts and invoices have similar content or layout, and training on their combination may boost the training over each of them individually. 
While for other two document types, \model~(H) has a slightly drop, 0.88\% on \bks~and 0.47\% on \bl~in \fone{} score.
This finding indicates that with our task formulation, a model can be trained on multiple document types and works well.
This may be very helpful for KIE over low-resource document types.

In addition, from~\tabref{field-result} we can see
\model~(H)~achieves significantly higher performance on the two ``column field'' (the value being a column of a table), \emph{Produce Code}  and \emph{Product Description}, on invoices than \model~(S), achieving 76.47\% for \emph{Produce Code} and 90.70\% for \emph{Product Description} in \fone{} score.
Through error analysis, we find that the \model~(S) model tends to predict a bigger region than the \model~(H), covering more content than the golden answer.
The underlying reasons need to be further investigated.
In \figref{col} , we present four predicted examples for extracting a column field \emph{Product Description}.

\subsection{Ablation Study}
We conduct ablation study w.r.t. the layout attention masking, layout attention bias, and Region Categorization module.
The results are summarized in \tabref{ablation}.

We first analyze the influence of our soft layout masking and bias mechanism. 
By removing either layout masking or layout bias, the model performance has a significantly drop on each of the four datasets. 
This indicates that they are both effective. 
Comparing with layout attention masking, the layout attention bias contributes more to the performance, bringing an increase in \fone{} score  of 5.1\%  on \sroie, 2.15\% on \inv, 5.62\% on \bks~and 7.68\% on \bl. 
Both layout attention masking and layout attention bias influence the performance most on \bl, gaining 3.19\% and 7.68\% \fone{} score respectively. 
This indicates they are more effective for bill of ladings compared to other document types.
A possible explanation would be, bill of ladings often have more complex layouts than other document types.

Then, we investigate the importance of the Region Categorization module.
By removing this module and only using region selection for deciding the target region, about 1\% decrease on EM or \fone{} can be observed.
This indicates the necessity of the multi-task training setting.
. 

\begin{table}[htb]
\centering
\footnotesize
\begin{tabular}{L{1.4cm}R{1.2cm}R{1.2cm}R{1.2cm}R{1.2cm}}
\toprule
\multirow{2}{*}{ \bf Model } & \bf \sroie & \bf \inv & \bf \bks & \bf \bl \\
\cmidrule{2-5}
& \bf EM/\fone{} & \bf EM/\fone{} & \bf EM/\fone{} & \bf EM/\fone{} \\
\midrule
Full \model &  90.98/95.32 &  77.96/90.68 &  87.17/90.70 &  74.80/81.68 \\
- LM & 87.78/94.94 & 77.20/90.23 & 85.12/88.98 & 71.86/78.49 \\
- LB & 82.77/90.22 & 75.48/88.53 & 81.40/85.08 & 66.72/74.00 \\
- RC & 88.58/95.64 & 77.03/90.28 & 86.12/89.70 & 74.71/80.46 \\

\bottomrule
\end{tabular}
\caption{Ablation results using \model~(H). The minus sign indicates removal of the mentioned part from the full \model.
\textit{LM}: Layout Masking;
\textit{LB}: Layout Bias;
\textit{RC}: Region Categorization;
}
 \label{tab:ablation}    
\end{table}

\section{Conclusion and Future Work}
We proposed a new formulation of the KIE task, namely treating it as a region prediction problem rather than the traditional sequence tagging.
This design greatly relieves the difficulty of processing non-flatten sequences that are very common in table-text documents, and meanwhile enables the model to be applied to various document types. 
In our experiments, we find that the proposed regions sometimes contain few even no valid regions, which makes our model difficult to be trained. 
Therefore, for future work we would like to upgrade the Region Proposal module to ensure there will be enough valid regions in the proposals.
%

\bibliography{aaai22}


\end{document}